%% file: MAIN.tex
\documentclass[10pt]{article} % For LaTeX2e
\usepackage[preprint]{tmlr}

\input{math_commands.tex}

\usepackage{hyperref}
\usepackage{url}
\usepackage{xspace}
\usepackage{xargs}
\usepackage{amsfonts}
\usepackage{booktabs}
\usepackage{ifthen}
\usepackage{float}
\usepackage{graphicx}
\usepackage{wrapfig}

\usepackage{amsmath}
\usepackage{colortbl}
\usepackage{subcaption}  % For subfigures (modern package for side-by-side plots)

\newcommand{\cort}{\citet{cornille2024learning}\xspace}

\newcommand{\actionspace}{\mathcal{A}}
\newcommand{\textspace}{\mathcal{X}}

\newboolean{negative_result_angle}

\title{End-to-end Planner Training for Language Modeling}

\author{
    \name Nathan Cornille\thanks{Corresponding author. Please direct correspondence to \texttt{nathan.e.d.cornille@gmail.com}.}\quad 
    \name Florian Mai\quad 
    \name Jingyuan Sun\quad 
    \name Marie-Francine Moens \\
    \\
    \addr Department of Computer Science, KU Leuven
}

\def\month{MM}  % Insert correct month for camera-ready version
\def\year{YYYY} % Insert correct year for camera-ready version
\def\openreview{\url{https://openreview.net/forum?id=XXXX}} % Insert correct link to OpenReview for camera-ready version

\begin{document}

\maketitle

\begin{abstract}
\iffalse\section{ABSTRACT}\fi
Through end-to-end training to predict the next token, LLMs have become valuable tools for various tasks. Enhancing their core training in language modeling can improve numerous downstream applications.
A successful approach to enhance language modeling uses a separate planning module to predict abstract labels of future sentences and conditions the LM on these predictions. However, this method is non-differentiable, preventing joint end-to-end tuning of the planner with the LM. We propose an effective method to improve this approach by enabling joint fine-tuning of the planner and the LM. We show that a naive way of approximating the gradient of selecting a label via the straight-through estimator is not effective. Instead, we propose to use the predicted label probabilities as mixing weights to condition the LM on a weighted average of label embeddings in a differentiable manner. 
This not only enables joint fine-tuning of the planner and the LM, but also allows the LM to draw on the full label distribution predicted by the planner, retaining more information.
Our experimental results show consistent improvements in perplexity.
\end{abstract}

\section{Introduction}

Large Language Models (LLMs) currently lay the foundation for excellent performance across a variety of downstream tasks.
They are not trained specifically for these tasks, but are pretrained in next token prediction, on trillions of tokens (typically followed by a short Reinforcement Learning from Human Feedback (RLHF) phase).
Improving the effectiveness of the language modeling phase can be expected to lead to greater usefulness on many downstream tasks.

\iffalse\subsection{Existing attempts + gaps}\fi

\cort are able to improve perplexity and generation quality by 1) pretraining a separate planning module to predict an abstract, discrete label of the next sentence (which they call "writing action", produced through unsupervised clustering) and 2) teaching the LM to condition on the planner's predictions while doing low-level next token prediction.
This enables factorizing the language modeling task into prediction of abstract, high-level information processing and concrete, low-level generation conditioned on the high-level plan.
Motivated by the idea of keeping the trained planner compatible with any LM, \citet{cornille2024learning} do not fine-tune the planner during step (2).
However, this means that the whole system is not trained end-to-end, breaking with one of the core concepts that make deep learning, and thus LLMs, so successful.

Therefore, the objective of this paper is to further improve LM performance by enabling end-to-end joint fine-tuning of the planner and the LM.
This comes with two challenges: \textbf{(i)} The discrete choice of an action is non-differentiable. \textbf{(ii)} Useful high-level features that the planner has learned during step (1) may be forgotten due to the low-level objective in step (2).

In order to address challenge \textbf{(i)}, we first investigate the straight-through estimator as a method for approximating the gradient of the hard-selection of a writing action. 
Finding that this is not effective, we propose a method that uses the planner-predicted action probabilities to compute a weighted average of the action embeddings, rather than selecting an embedding based on which action has the highest probability. 
This has two advantages: 
First, it allows us to compute an exact gradient. 
Second, it does not discard valuable information contained in the label distribution predicted by the planner.
We perform probing experiments that corroborate the hypothesis that using the full range of planner-predicted probabilities results in representations that are more informative about distant tokens.

To address challenge \textbf{(ii)}, we show two approaches that succeed at preventing catastrophic forgetting: one approach is to keep the planner parameters frozen for the first half of training before unfreezing them for the second half of training, another approach is to fine-tune the planner with a mix of its original Next-Action Prediction objective and the Next-Token Prediction objective.
Moreover, we show that pretraining the planner with only the low-level objective (Next Token Prediction) damages performance, again underscoring the importance of the high-level objective (Next Action Prediction).

% While end-to-end training proves effective at improving perplexity, this improvement does not translate consistently into better generation metrics.
As observed in prior work, generation metrics can be improved by training the language model with oracle actions, as this means the language model learns to rely on the plan more strongly. 
However, this teacher forcing damages perplexity due to the mismatch of oracle actions during training with imperfect planner-predicted actions during evaluation, known as exposure bias.
We show that we can strike a balance in this trade-off by mixing oracle and planner-predicted actions with a certain probability during training.

\iffalse\subsection{Contributions summary}\fi
Our contributions can be summarized as follows:
\begin{itemize}
    \item We propose a method that addresses the challenges of joint end-to-end training of a high-level planner module and a low-level language model.
    \item We compare our method using two LM backbones (GPT-2-small and OLMo-1B), demonstrating improvements in perplexity.
    \item We show that improvements depend on two key factors: (i) avoiding catastrophic forgetting of the planner's high-level knowledge—achieved by delaying the unfreezing of planner parameters or including the planner's high-level objective during training; and (ii) ensuring the LM has access to all planner-predicted probabilities rather than just the top prediction.
    \item We identify a trade-off between perplexity and generation quality that depends on the use of oracle or planner-predicted actions during training, and demonstrate that mixing oracle and planner-predicted actions balances this trade-off effectively. 
\end{itemize}

\section{Related Work}

Our method is related to four fields of research.
1) Conditioning the language model on a writing action is akin to \emph{Controllable Text Generation}, where a language model is conditioned on attributes of various types, such as content or style, to guide generation.
%From one perspective, we can consider the planning module as something external that provides the LM with label-predictions to condition on, the planning module is related to other works that seeks to condition language models on plans\footnote{A `plan' here can be broadly interpreted as some abstract property that the text about to be written should correspond to.} of some kind.
2) Since the planner operates at a higher level than the LM, our method is related to \emph{Hierarchical Language Modeling}.
%From a second perspective, because this work adds a training phase to the planner in which it is trained jointly with the LM, it is also closely related to Hierarchical Language Modeling.
3) As the core contribution of our paper is enabling end-to-end training, we discuss techniques for \emph{bridging the differentiability gap}. 
% A key component of our contribution is to enable the planner to be optimized jointly with the LM.

\paragraph{Conditioning language models on plans}

In Controllable Text Generation~\citep{zhang2023survey}, human-provided inputs are used to condition text generation, but we use latent, planner-produced inputs rather than human-provided ones.
Quiet-STaR~\citep{zelikman2024quiet} conditions next token prediction on a model-produced input, but this is applied at every token, making it more computationally intensive, and is in text space rather than latent space. 
\citet{wang_guiding_2023} use planning tokens with a similar objective as the writing actions we consider, but they use an \textit{internal} planner, i.e., the LM itself, to predict planning tokens, in contrast to the external planner from \cort, who found an internal planner to be less flexible and less effective.
%. The
%an internal planner is less flexible, and was found to be less effective in \cort.

\paragraph{Hierarchical Language Modeling}

\citet{chung_hierarchical_2017}, \citet{li2022learning}, and \citet{subramanian2020multi} all factorize language modeling into a per-token part and a slower `per text-unit' part (e.g., per-phrase, per-sentence, or every fixed number of tokens). The per text-unit component in their work is only optimized for predicting concrete tokens. 
The key difference with this work is that our per text-unit component (i.e., the planner) is pretrained to predict targets in an abstract sentence space, and only then fine-tuned together with the per-token component (i.e., the LM) to predict concrete tokens.
\citet{marfurt2021sentence} and \citet{jhamtani2020narrative} also aim to improve language modeling by decomposing at the sentence and word level, but unlike \cort and this work do not operate at an \textit{abstract} sentence level.

\paragraph{Bridging differentiability gap}
A key component of our contribution is to enable the planner to be optimized jointly with the LM.
A number of methods exist aimed at estimating a gradient for non-differentiable operations such as an argmax. Policy Gradient methods are one type of approach 
% (also known as score-function gradient estimators~\citep{huijben2022review}), 
which are common in reinforcement learning~\citep{williams1992simple,sutton1999policy,greensmith2004variance,schulman2017proximal}. While unbiased, these tend to have high variance.
Relaxed gradient estimators~\citep{maddison2016concrete,paulus2020gradient} constitute another approach, they replace a discrete sample with a continuous variable to calculate the gradient. An important variant of these is the straight-through estimator~\citep{bengio2013estimating}, which uses a hard max for the forward pass, but a softmax for the backward pass. 
While this is a popular biased, but low-variance estimator of the gradient, we demonstrate that it yields unsatisfactory results in our scenario.

% In our work, one of our insights is that there is in fact no inherent need to get a discrete sample from the planner.
% The planner-predicted probabilities can simply be used as mixing weights to get a weighted average of action embeddings, which allows us to calculate an exact gradient.

\paragraph{Exposure bias and teacher forcing}
The method proposed by \citet{cornille2024learning} learns a \emph{cascaded model}, where the predicted class of the first model A (the planner) is propagated into the second model B (the language model).
An old technique in such cases is to replace the actual prediction of model A with the ground truth signal \citep{jordan1986attractor, pineda1988dynamics}.
In sequence learning networks such as RNN this is known as \emph{teacher forcing}, which can help stabilize training~\citep{williams1989learning}.
However, on the flip side, this approach can introduce an \emph{exposure bias}, in which the distribution of inputs seen by model B at training time differs from inference time, leading to sub-optimal results~\citep{bengio2015scheduled, lamb2016professor}.
There are several ways to circumvent the exposure bias problem. The first type of approach aims to mitigate the training and test distribution mismatch. Scheduled sampling~\citep{bengio2015scheduled} selects the self-predicted input with probability $p$ and the ground truth input with probability $1 - p$ during training. $p$ is scheduled to increase from $0$ to $1$ over the course of training to retain the best from both worlds. Professor Teaching~\citep{lamb2016professor} uses an adversarial method to make the train and test distribution similar.
Another type of approach circumvents the problem altogether by training the model end-to-end on the metric of interest, e.g. \citet{ranzato2015sequence} use self-predicted outputs in combination with reinforcement learning for text generation, and \citet{graves2014towards} directly optimizes for low word error rate in speech recognition.
In this paper, we explore both scheduled sampling and end-to-end training to deal with the exposure bias.

\section{Methodology}
\label{sec:meth}
% We build on the approach proposed in \cort, but introduce a significant modification in how we fine-tune the Language Model (LM) conditioned on predicted writing actions. 
% For a comprehensive understanding of the existing framework, we refer to the methodology section of the prior work. 
% Here, we will briefly summarize the common elements and then focus on the novel aspects of our approach.
We summarize the common elements we preserve in \ref{subsec:common}. We then detail the prior way of interfacing the planner and the language model in \ref{subsec:prior_meth}, and our novel way in \ref{subsec:new_meth}.

\subsection{Overview of Common Methodology}
\label{subsec:common}
We consider the task of language modeling, where the goal is to estimate the probability $p(x_1 x_2 \dots x_n) $ for any text sequence \( X = x_1 \dots x_n \in \textspace \). 
We also refer to this task as \textit{Next Token Prediction}.
The probability is factorized into the product of probabilities of each token given the preceding tokens: \( p(x_1 \dots x_n) = \prod_{i=1}^n p(x_i | x_1 \dots x_{i-1}) \). Unlike a standard LM, the predicted probability in our method is conditioned on additional
%writing actions \( a \in \actionspace \), which are predicted by an external planner module \( \planner : \textspace \rightarrow \actionspace \).
predictions by an external planner.

To obtain training data for this planner, each sentence \( t_j \) in the corpus is first embedded into a low-dimensional vector \( \mathbf{z}_j \) using some pretrained text encoder (e.g., Sentence-BERT). 
K-means clustering on all embedded text units from the corpus yields clusters that are used as abstract labels (or `writing actions')  \( a \in \actionspace \).
A cluster's centroid serves as the (initial) action embedding.

%In order to get this planner, each text unit\footnote{Text units correspond to sentences in \cort and this paper.} \( t_j \)  is first embedded into a low-dimensional vector \( \mathbf{z}_j \) using some pretrained text encoder (e.g., Sentence-BERT). After applying K-means clustering on these embedded text units, the index of the nearest cluster of each text unit is used as an abstract label or `writing action'  \( a \in \actionspace \).
The planner module is then pretrained to predict the writing action $a_i$ that corresponds to sentence $t_i$ based on the context of the preceding text units $t_1, \dots, t_{i-1}$. This task is called \emph{Next Action Prediction}.
The planner module is based on a custom Transformer architecture that first embeds each sentence independently into a single vector and then contextualizes them with a Transformer encoder. For details, please refer to \cort.

\subsection{Prior approach to Language Model Fine-tuning}
\label{subsec:prior_meth}
In \cort, the input \( X \) is split into $m$ sentences \( X = t_1 \dots t_m \), where each sentence \( t_j = x^j_1 \dots x^j_{n_j} \) has a single associated \textit{predicted} writing action \( \hat{a}_j \), predicted by a pretrained planner.
The fine-tuning objective then is to estimate \( \prod_{j=1}^m \prod_{i=1}^{n_j} p(x_i^j | \hat{a}_1 \dots \hat{a}_j, t_1 \dots t_{j-1} x_1^j \dots x_{i-1}^j) \).

The predicted actions are one-hot encoded, and used as index to select action embeddings in the action embedding matrices $E_A^l$ that are added at various layers $l$ of the LM, resulting in a vector $r^l_j = E_A^l(\hat{a}_j)$. That vector is then projected and element-wise added to each hidden representation in layer $l$, just after the multiplication with attention weights and just before the output projection, similar to Llama-Adapter~\citep{zhang2023llama}.

Importantly, planner parameters are not updated during LM fine-tuning, disallowing the planner to improve its action predictions beyond matching cluster labels, by tailoring them to the LM.

\subsection{Novel Joint Planner-Language Model Fine-tuning}
\label{subsec:new_meth}

We hypothesize that the planner can enhance the utility of its predictions by being fine-tuned for next-token prediction, jointly with the LM, after being pretrained for Next Action Prediction. 
Hence, we want to enable the gradient to be passed into the planner.

A naive way to achieve this is using a straight-through estimator, which involves using a hard max (Eq. \ref{eq:hardmax}) for the forward pass but a softmax (Eq. \ref{eq:softmax}) for the backward pass: 
\begin{equation}
\label{eq:hardmax}
    \operatorname{onehot}(\operatorname{argmax}(\mathbf{s}))
\end{equation}

\begin{equation}
\label{eq:softmax}
    \left[ \frac{e^{s_1}}{\sum_{j=1}^{|A|} e^{s_j}}, \frac{e^{s_2}}{\sum_{j=1}^{|A|} e^{s_j}}, \ldots, \frac{e^{s_{|A|}}}{\sum_{j=1}^{|A|} e^{s_j}} \right]
\end{equation}
Here, $\mathbf{s} = \left[s_1, \ldots, s_{|A|} \right] $ is the vector of planner-predicted logits for each of $|A|$ possible actions.

However, the straight-through estimator is limited by being a biased estimator of the gradient.
A more effective method arises from the insight that there is no inherent need to select only one action. Instead, we can use the planner probabilities to obtain a weighted average of the action embedding columns:

%\[
\begin{equation}\label{eq:soft-selection}
\mathbf{r}_j^l = \sum_{a \in \actionspace} \operatorname{softmax}(\mathbf{s})_a \cdot E_A^l(a),    
\end{equation}

%\]

\begin{figure}[ht]
    \centering
    \includegraphics[width=0.8\textwidth]{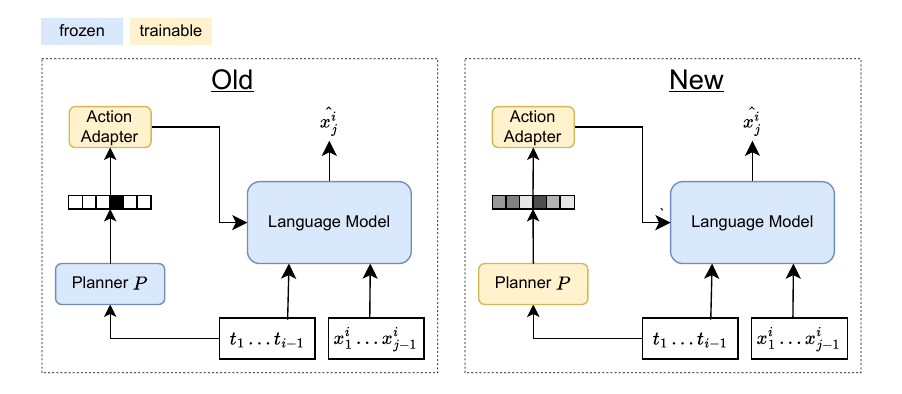}
    \caption{Illustration of our proposed improvement. The planner predicts a distribution over actions, which is used as mixing weights to compute a weighted average of the action embeddings. 
    This allows the planner to be fine-tuned jointly with the LM.}
    \label{fig:mainfig}
\end{figure}

Figure~\ref{fig:mainfig} illustrates this.

This approach not only provides an exact gradient but also allows the LM to make use of the full set of probabilities assigned by the planner to each writing action, rather than only the most probable action.

\paragraph{Preventing catastrophic forgetting}
Unfreezing the planner immediately along with the LM poses the risk of catastrophic forgetting~\citep{french1999catastrophic} of the learnings from the Next-Action Prediction pretraining stage.
One common mitigation strategy is to unfreeze only after some training steps~\citep{howard2018universal}, hence we evaluate a setting in which we unfreeze the planner only halfway through training.
Another strategy we evaluate is to fine-tune the planner with a mix of its original Next-Action Prediction objective, and the Next-Token Prediction objective.

\section{Experimental Setup}
% The objectives of our experiments are to 
The purpose of our experiments is twofold.
First, we want to test our hypothesis that end-to-end joint planner-LM training is beneficial for language modeling performance.
Second, we want to validate the design decisions we made to enable end-to-end training: a) Using a soft-selection via weighted average rather than a hard selection, and b) mitigating catastrophic forgetting by unfreezing the planner only after half the training or using a mixed Next Action / Next Token prediction objective.

\subsection{Dataset and backbone models}

We train and evaluate our models on the same dataset as \cort, i.e., subsets of English Wikipedia articles from the ``20220301.en'' version from Huggingface\footnote{\url{https://huggingface.co/datasets/wikipedia}
%, Creative Commons Attribution Share Alike 3.0
}.
Wikipedia articles have the advantage that they cover an extensive range of topics, while also being structured in a way that makes them well-suited for leveraging an abstract planner.
We perform experiments both using the small GPT2 model~\citep{radford2019language} and OLMo 1B~\citep{groeneveld2024olmo} as starting points for the LM.
Model details and hyperparameters are provided in Appendix~\ref{app:hyperparameters}.

\subsection{Evaluation}
% Our main metric is perplexity, which is the default metric used for language modeling, corresponding to the inverse geometric mean of the probability of true texts according to the language model.
% For completeness, we also report the edit distance metric reported in \cort, which measures how well model-generated text corresponds to ground-truth text in terms of writing action sequences, though we expect the improvement of fine-tuning the planner for perplexity to be mostly in perplexity.

\paragraph{Primary evaluation}
Our main metric is perplexity, which is the default metric used for language modeling, corresponding to the inverse geometric mean of the probability of true texts according to the language model.

\paragraph{Generation evaluation}
As in \cort, we complement perplexity, which does not directly assess generated text, with generation metrics.
We report ROUGE-2 (F1)~\citep{lin2004rouge} and MAUVE~\citep{pillutla2021mauve} to evaluate generated texts at the surface level, and Levenshtein distance~\citep{levenshtein1966binary} and latent perplexity~\citep{deng2022model} to assess text quality at an abstract level.
For the surface level, ROUGE-2 evaluates bigram overlap between generated and real text, while MAUVE measures the divergence between model and true data distributions by comparing generated and real texts unconditionally.
For the abstract level, we first map true and generated texts onto the sequence of writing actions that correspond to them.
Levenshtein distance then measures the edit distance between generated and ground-truth writing action sequences, and latent perplexity estimates how well the generated sequence aligns with a latent HMM-based critic trained on real texts.
We refer to appendix \ref{app:geneval} for more details about the generation evaluation.

\paragraph{Probing}

In order to understand how the different training setups influence what information the model (un)learns, we use probing classifiers on top of the (frozen) representations to determine how well they predict the upcoming tokens. The choice of probing classifier is not straight-forward~\citep{belinkov2022probing}. We choose linear probing classifiers to measure to what extent the information about upcoming tokens can be easily extracted (i.e., is linearly separable) from the representations, rather than be learned by the probe itself~\citep{alain2016understanding}.

% TODO: Information about different types of probing locations
We probe representations at two kinds of locations inside the model. 
First, we probe the output from the action embedding inside the adapter, which contains information only from the planner (Pre-merge). 
Second, we probe the representation after the information of the planner has been mixed with the contextualized information from the LM itself (Post-merge). 
We train probes at every layer where the planner information is infused.

\begin{wrapfigure}{r}{0.5\textwidth}
    \setlength{\belowcaptionskip}{-35pt}  % Adjust the amount as needed
    \centering
    \includegraphics[width=0.48\textwidth]{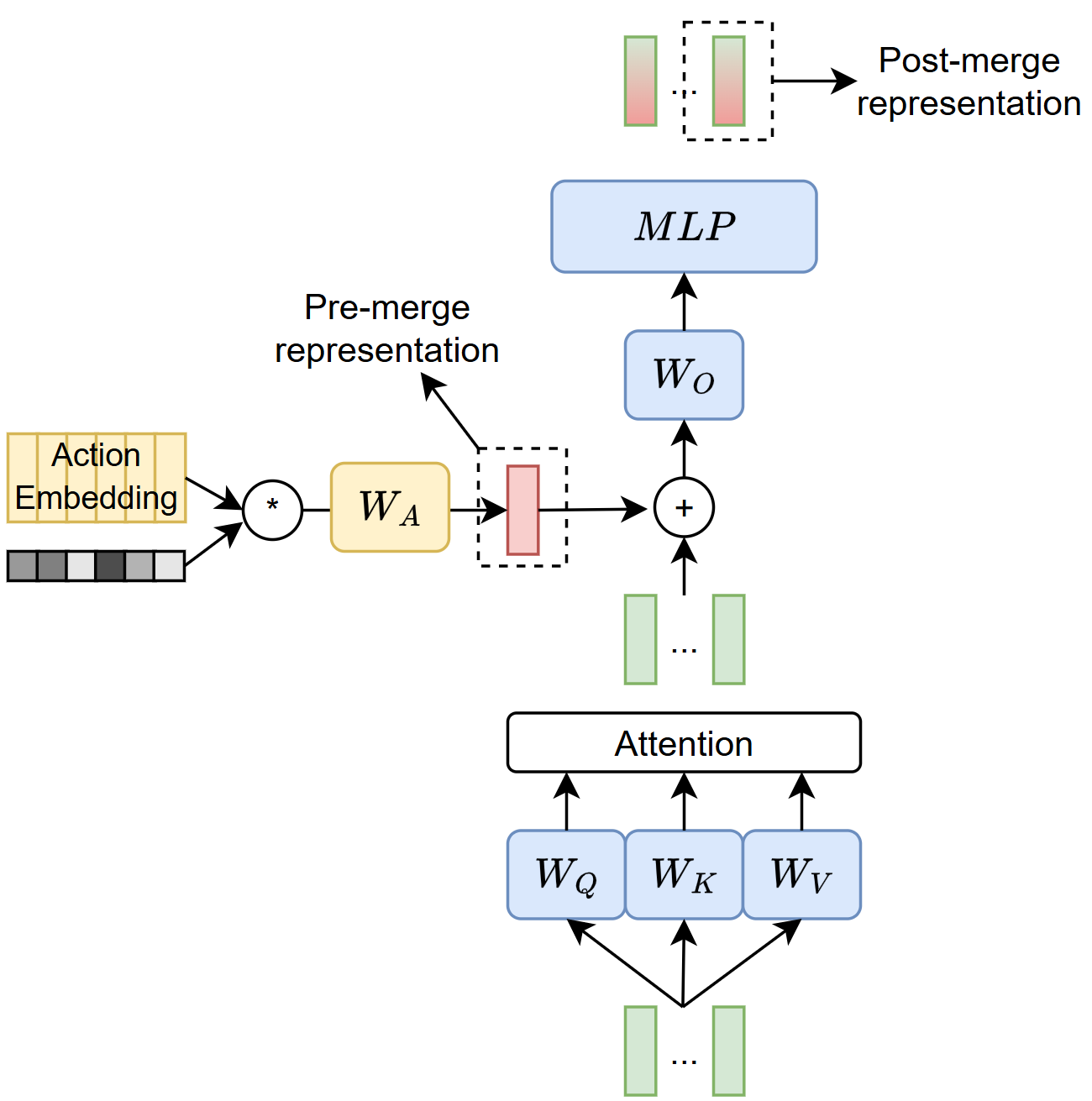}
    \caption{Illustration of the probing locations inside the model.}
    \label{fig:probe_locations}
\end{wrapfigure}

Figure \ref{fig:probe_locations} illustrates this.

\subsection{Settings}

% Our methods differ in two regards: 1) Hard-selection via the straight-through estimator (Eq.~\ref{eq:hardmax} and \ref{eq:soft-selection}) or soft-selection via weighted averages (Eq.~\ref{eq:soft-selection}). 2) Whether the planner's parameters are unfrozen immediately (\emph{Unfrz immediate}), halfway through training (\emph{Unfrz halfway}), or never (\emph{Unfrz never}).
% %
% In order to test the benefit of end-to-end training, we compare the \cort baseline to our best method \emph{Unfrz halfway}.
% To validate our method design decisions, we compare the respective best results in each setting.
% To rule out that the benefit of soft-mixing is not merely due to mixing multiple actions, we train a variant of the soft-selection method that always applies uniform weighting across all actions (\emph{Uniform}).

\paragraph{Variations of end-to-end planner}
We evaluate the impact of 4 properties of the end-to-end planner.

First, whether the planner's parameters are unfrozen immediately (\emph{Unfrz immediate}), halfway through training (\emph{Unfrz halfway}), or never (\emph{Unfrz never}).
We expect that immediately unfreezing the planner when the LM hasn't adapted to it yet might lead to catastrophic forgetting, while not unfreezing it at all doesn't allow the planner to tune itself to the LM.

Second, similarly aimed at preventing catastrophic forgetting, we evaluate the effect of continuing to train the planner for its Next-Action Prediction objective at the same time as also tuning end-to-end for Next-Token Prediction Objective.

Third, we evaluate the effect of replacing soft-selection via weighted averages (Eq.~\ref{eq:soft-selection}) with hard-selection via the straight-through estimator (Eq.~\ref{eq:hardmax} and \ref{eq:soft-selection}).

Finally, because we are now able to train the planner end-to-end, we evaluate whether its Next-Action-Prediction (NAP) pretraining objective is still necessary by assessing models in which the planner is only pretrained with a Next-Token-Prediction objective.
Specifically, we replace NAP training of the planner with an end-to-end stage in which we keep the LM parameters frozen.

\paragraph{Baselines}

To rule out that the benefit of soft-mixing is not merely due to mixing multiple actions, we train a variant of the soft-selection method that always applies uniform weighting across all actions (\emph{Uniform}).

Our main baselines are the planner models proposed in \cort. 
They have two variants: one pretrained on oracle actions (OA), and one pretrained on predicted actions (PA).
\cort observed a trade-off between these variants: while PA had better perplexity, OA performed better in some generation metrics.
We explore this trade-off more in-depth by making models that mix oracle-action and predicted actions during pretraining.

\section{Results and Discussion}
\subsection{Main results}
\iffalse\subsubsection{Table}\fi
Table \ref{tab:main_results} shows our main results.
\begin{table}[ht!]
\centering
\small
\scalebox{0.7}{
\begin{tabular}{@{}l|ccccc|ccccc@{}}
\toprule
Base LM & \multicolumn{5}{c}{GPT2} & \multicolumn{5}{c}{OLMO} \\
\cmidrule{1-11}
 & PPL $\downarrow$ & MAUVE $\uparrow$ & Latent PPL $\downarrow$ & ROUGE-2 $\uparrow$ & Edit $\downarrow$ & PPL $\downarrow$ & MAUVE $\uparrow$ & Latent PPL $\downarrow$ & ROUGE-2 $\uparrow$ & Edit $\downarrow$ \\
Setting &  &  &  &  &  &  &  &  &  &  \\
\midrule
\cmidrule{1-11}
\multicolumn{11}{c}{\textbf{Baselines}} \\
\cmidrule{1-11}
\textit{\cort OA} & \cellcolor[HTML]{f14432} 26.94 & \cellcolor[HTML]{e3eef9} 0.447 & \cellcolor[HTML]{f7fbff} 91.60 & \cellcolor[HTML]{f7fbff} 0.0193 & \cellcolor[HTML]{f2f8fd} 3.69 & \cellcolor[HTML]{f14432} 11.99 & \cellcolor[HTML]{61a7d2} 0.411 & \cellcolor[HTML]{f7fbff} 76.20 & \cellcolor[HTML]{f7fbff} 0.0278 & \cellcolor[HTML]{e7f0fa} 3.26 \\
\textit{\cort PA} & \cellcolor[HTML]{fee3d6} 25.55 & \cellcolor[HTML]{d4e4f4} 0.435 & \cellcolor[HTML]{c3daee} 205.90 & \cellcolor[HTML]{94c4df} 0.0169 & \cellcolor[HTML]{c3daee} 3.78 & \cellcolor[HTML]{fee7db} 11.46 & \cellcolor[HTML]{f2f7fd} 0.563 & \cellcolor[HTML]{aacfe5} 178.20 & \cellcolor[HTML]{d7e6f5} 0.0253 & \cellcolor[HTML]{cadef0} 3.31 \\
\textit{Uniform} & \cellcolor[HTML]{fa6547} 26.69 & \cellcolor[HTML]{4a98c9} 0.378 & \cellcolor[HTML]{4a98c9} 354.27 & \cellcolor[HTML]{4a98c9} 0.0159 & \cellcolor[HTML]{4a98c9} 3.91 & \cellcolor[HTML]{fc8161} 11.81 & \cellcolor[HTML]{4a98c9} 0.396 & \cellcolor[HTML]{4d99ca} 256.13 & \cellcolor[HTML]{81badb} 0.0219 & \cellcolor[HTML]{539ecd} 3.43 \\
\cmidrule{1-11}
\multicolumn{11}{c}{\textbf{Ours (soft-selection)}} \\
\cmidrule{1-11}
\textit{Unfrz immediate} & \cellcolor[HTML]{feeae0} 25.42 & \cellcolor[HTML]{c4daee} 0.423 & \cellcolor[HTML]{a6cee4} 245.48 & \cellcolor[HTML]{c9ddf0} 0.0178 & \cellcolor[HTML]{f7fbff} 3.68 & \cellcolor[HTML]{ffede5} 11.42 & \cellcolor[HTML]{f2f7fd} 0.564 & \cellcolor[HTML]{a0cbe2} 187.61 & \cellcolor[HTML]{eff6fc} 0.0271 & \cellcolor[HTML]{bad6eb} 3.33 \\
\textit{Unfrz halfway }& \cellcolor[HTML]{fff5f0} 25.23 & \cellcolor[HTML]{c2d9ee} 0.422 & \cellcolor[HTML]{c3daee} 205.14 & \cellcolor[HTML]{d9e7f5} 0.0183 & \cellcolor[HTML]{b2d2e8} 3.80 & \cellcolor[HTML]{fff5f0} 11.37 & \cellcolor[HTML]{ebf3fb} 0.551 & \cellcolor[HTML]{bad6eb} 163.78 & \cellcolor[HTML]{eef5fc} 0.0270 & \cellcolor[HTML]{f7fbff} 3.23 \\
\textit{Unfrz never} & \cellcolor[HTML]{fff0e8} 25.32 & \cellcolor[HTML]{bed8ec} 0.420 & \cellcolor[HTML]{ccdff1} 187.54 & \cellcolor[HTML]{bad6eb} 0.0175 & \cellcolor[HTML]{d8e7f5} 3.74 & \cellcolor[HTML]{fee2d5} 11.49 & \cellcolor[HTML]{e9f2fa} 0.546 & \cellcolor[HTML]{bad6eb} 163.81 & \cellcolor[HTML]{eff6fc} 0.0271 & \cellcolor[HTML]{d6e5f4} 3.29 \\
\cmidrule{1-11}
\multicolumn{11}{c}{\textbf{Ours (straight-through)}} \\
\cmidrule{1-11}
\textit{Unfrz immediate}& \cellcolor[HTML]{fcbca2} 25.94 & \cellcolor[HTML]{8dc1dd} 0.401 & \cellcolor[HTML]{87bddc} 281.34 & \cellcolor[HTML]{5fa6d1} 0.0162 & \cellcolor[HTML]{6dafd7} 3.87 & \cellcolor[HTML]{fed8c7} 11.53 & \cellcolor[HTML]{eaf2fb} 0.548 & \cellcolor[HTML]{85bcdc} 208.12 & \cellcolor[HTML]{a5cde3} 0.0229 & \cellcolor[HTML]{9fcae1} 3.36 \\
\textit{Unfrz halfway} & \cellcolor[HTML]{fed9c9} 25.66 & \cellcolor[HTML]{f7fbff} 0.464 & \cellcolor[HTML]{b2d2e8} 230.00 & \cellcolor[HTML]{a3cce3} 0.0171 & \cellcolor[HTML]{bad6eb} 3.79 & \cellcolor[HTML]{ffede5} 11.42 & \cellcolor[HTML]{eaf2fb} 0.547 & \cellcolor[HTML]{a3cce3} 185.76 & \cellcolor[HTML]{d8e7f5} 0.0254 & \cellcolor[HTML]{b2d2e8} 3.34 \\
\cmidrule{1-11}
\multicolumn{11}{c}{\textbf{Ours (NAP during fine-tuning)}} \\
\cmidrule{1-11}
\textit{Unfrz immediate} & \cellcolor[HTML]{fff5f0} 25.24 & \cellcolor[HTML]{f2f7fd} 0.459 & \cellcolor[HTML]{d0e2f2} 177.34 & \cellcolor[HTML]{a9cfe5} 0.0172 & \cellcolor[HTML]{e3eef8} 3.72 & \cellcolor[HTML]{ffede5} 11.42 & \cellcolor[HTML]{f7fbff} 0.576 & \cellcolor[HTML]{c3daee} 155.38 & \cellcolor[HTML]{e8f1fa} 0.0266 & \cellcolor[HTML]{d6e5f4} 3.29 \\
\cmidrule{1-11}
\multicolumn{11}{c}{\textbf{Ours (no NAP pretraining)}} \\
\cmidrule{1-11}
\textit{Unfrz immediate} & \cellcolor[HTML]{fdcab5} 25.80 & \cellcolor[HTML]{deebf7} 0.443 & \cellcolor[HTML]{8fc2de} 271.71 & \cellcolor[HTML]{85bcdc} 0.0167 & \cellcolor[HTML]{97c6df} 3.83 & \cellcolor[HTML]{fca78b} 11.69 & \cellcolor[HTML]{f1f7fd} 0.562 & \cellcolor[HTML]{6caed6} 227.46 & \cellcolor[HTML]{aacfe5} 0.0231 & \cellcolor[HTML]{4a98c9} 3.44 \\
\textit{Unfrz halfway} & \cellcolor[HTML]{fdc9b3} 25.82 & \cellcolor[HTML]{82bbdb} 0.397 & \cellcolor[HTML]{75b4d8} 299.62 & \cellcolor[HTML]{75b4d8} 0.0165 & \cellcolor[HTML]{c3daee} 3.78 & \cellcolor[HTML]{fcb296} 11.66 & \cellcolor[HTML]{e3eef8} 0.534 & \cellcolor[HTML]{6fb0d7} 224.99 & \cellcolor[HTML]{7db8da} 0.0218 & \cellcolor[HTML]{94c4df} 3.37 \\
\textit{Unfrz never} & \cellcolor[HTML]{fca486} 26.15 & \cellcolor[HTML]{d4e4f4} 0.435 & \cellcolor[HTML]{75b4d8} 299.51 & \cellcolor[HTML]{65aad4} 0.0163 & \cellcolor[HTML]{c3daee} 3.78 & \cellcolor[HTML]{fc8565} 11.80 & \cellcolor[HTML]{549fcd} 0.403 & \cellcolor[HTML]{4a98c9} 258.24 & \cellcolor[HTML]{4a98c9} 0.0204 & \cellcolor[HTML]{4a98c9} 3.44 \\
\bottomrule
\end{tabular}

}
\caption{Perplexity and generation metrics under different training and conditioning scenarios. Cells shaded in red show perplexity, those in blue show the generation metrics. A darker color indicates a worse result.}
\label{tab:main_results}
\end{table}

\iffalse\subsubsection{Table discussion}\fi
\paragraph{Benefit of end-to-end training}
Comparing \textbf{Ours (soft-selection)} to the \textbf{Baselines}, the results confirm our hypothesis that end-to-end joint planner-LM training can improve language modeling performance compared to the prior approach, with our best setting improving by 0.3 (GPT-2) and 0.08 perplexity (OLMo), respectively over \textit{\cort PA} (Predicted Actions).
% As expected, this perplexity improvement does not always translate into an improved edit distance, which we attribute to directly optimizing for perplexity.

We observe that this perplexity improvement does not always translate into improved generation metrics.
As noted in \cort, there is a trade-off between perplexity and performance on generation metrics stemming from the use of teacher forcing for the actions.
We examine this trade-off in more detail in section \ref{sec:tradeoff}.

\paragraph{Soft selection beats hard selection}
Comparing \textbf{Ours (soft-selection)} to \textbf{Ours (straight-through)}, we see that soft-selection variants are consistently better than straight-through variants.
This can be explained by two factors. 
First, the biased gradient estimates of the straight-through estimator might lead the learning astray. 
Second, soft-selection has the benefit of allowing the LM to draw on the full label distribution: In fact, the soft-selection \textit{Unfrz never} result shows that this alone is already beneficial, even without updating the planner.
This explanation is corroborated by the probing results presented in section~\ref{sec:probe}, which show that linear probes trained on the soft-selected planner output (rather than the hard-selected one) are better able to predict distant tokens.

Soft-selection also activates the full action embedding matrix at every prediction. However, the fact that \emph{Uniform} performs considerably worse shows that just using the full embedding matrix is \textit{not} responsible for the improvement.

\paragraph{Timing matters for planner unfreezing}
Keeping the planner frozen during part of the training is more effective than either immediately unfreezing the planner or keeping it frozen the entire time.
This is in line with our hypothesis that immediately unfreezing the planner leads to big initial gradients that erase some of the useful knowledge built up during the Next-Action-Prediction planner pretraining phase. 
On the other hand, not unfreezing the planner at all prevents the planner parameters from specializing toward perplexity minimization.

Our alternative approach to preventing catastrophic forgetting (\textbf{Ours (NAP during fine-tuning)}) achieves performance nearly on par with unfreezing the planner halfway.

\paragraph{Next-Action-Prediction objective cannot be left out completely}
Finally, we see that the models we run with \textbf{no NAP pretrainnig} are generally worse for both perplexity and generation metrics than \textbf{Ours (soft-selection)}.
This indicates that the abstract pretraining objective of the planner is still required, even when end-to-end training is possible.

\subsection{What do the models learn?}
\label{sec:probe}

\begin{figure}[ht]
    \centering
    % First plot
    \begin{subfigure}[b]{0.594\textwidth}
        \centering
        \includegraphics[width=\textwidth]{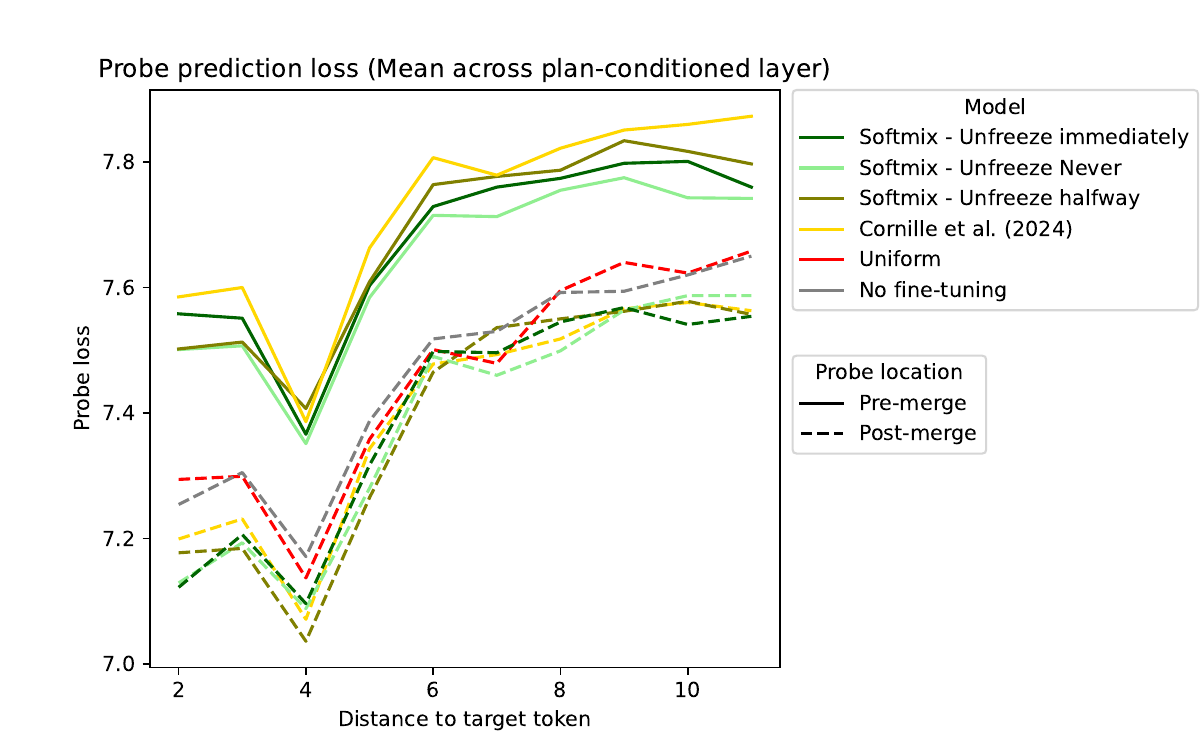}
        \caption{Probe Layer Mean}
        \label{fig:probe-perf-by-token}
    \end{subfigure}
    % Second plot
    \begin{subfigure}[b]{0.386\textwidth}
        \centering
        \includegraphics[width=\textwidth]{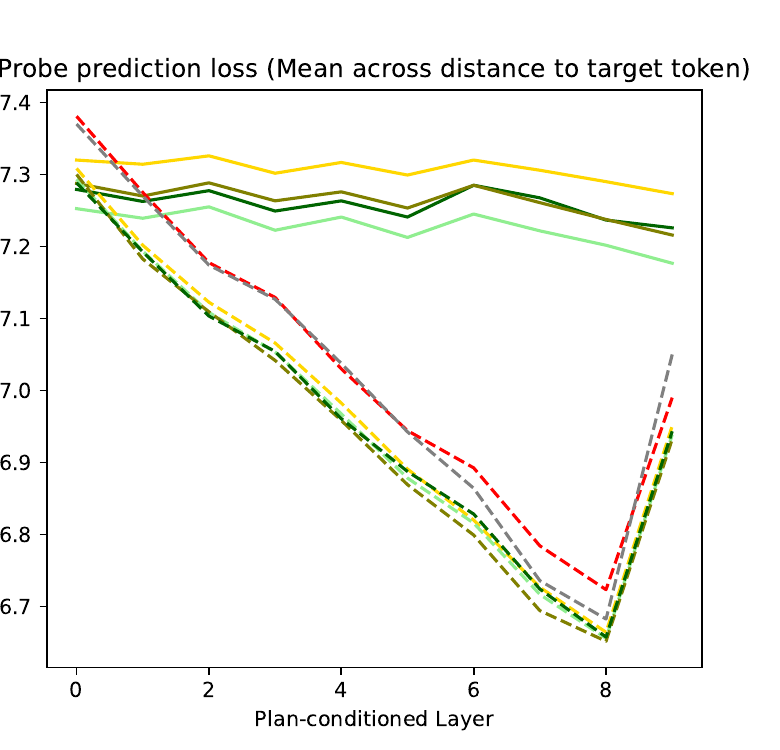}
        \caption{Probe Target Mean}
        \label{fig:probe-perf-by-layer}
    \end{subfigure}
    
    \caption{Plots showing probing performance at different layers and for different distances to the probe's target token.}
    \label{fig:probes}
\end{figure}

Figure~\ref{fig:probe-perf-by-token} shows the results of our probing experiments by distance to the target token.
Unsurprisingly, tokens farther away tend to be more difficult to predict.
% pre-merge fpjpl vs cornille et al.
Regarding pre-merge representations,
the \citet{cornille2024learning} baseline is notably worse than our proposed methods using softmix representations, which benefit from making better use of the full planner's predicted scores rather than only the argmax.  
% post-merge better than pre-merge
Generally, the post-merge representations are significantly better than the pre-merge representations. In fact, the language model alone, without any (useful) planner information ("Uniform") already predicts farther into the future than just the next token.
% adding pre-merge representation helps the performance.
However, adding the pre-merge representations of the planner yields further improvements. 
Moreover, freezing the pretrained planner at least for half the training epoch tends to retain more information about the upcoming tokens than unfreezing it immediately.
While this probing experiment cannot prove a causal mechanism, it is plausible that the improved performance observed in Table~\ref{tab:main_results} is at least partially attributable to the models ability to being better at predicting several tokens ahead.

Figure~\ref{fig:probe-perf-by-layer} shows that the information contained in pre-merge representations is largely independent of the layer, which is explained by the fact that lower layer representations do not feed into higher layer representations.
In contrast, post-merge results clearly show that higher layers, which are located closer to the output layer that performs the final token prediction, contain more information about the upcoming tokens.

\subsection{Trade-off between Perplexity and Generation Metrics}
\label{sec:tradeoff}
\begin{figure}[ht]
    \centering
    \includegraphics[width=.9\linewidth]{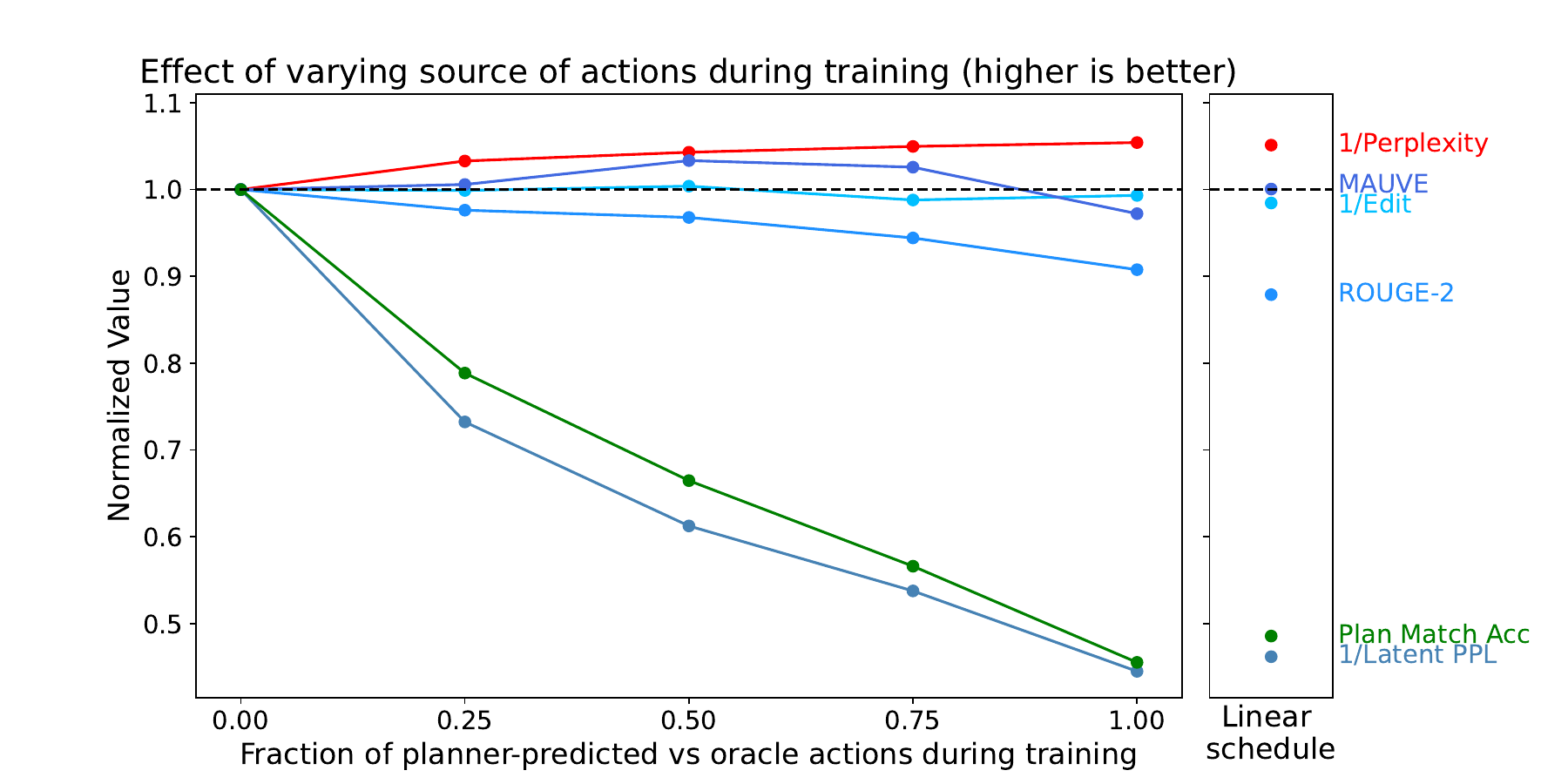}
    \caption{Relative improvement/worsening of our metrics as we increase the fraction of planner-predicted actions from zero (equivalent to \cort OA) to one (equivalent to \cort PA).
    Some metrics are inverted, so that higher is better for all metrics.}
    \label{fig:cms}
\end{figure}

% Setup
To investigate the trade-off between perplexity and generation metrics (MAUVE, ROUGE-2, Edit distance and Latent Perplexity), we train models that use a mixture of oracle and planner-predicted actions during training, where we vary the fraction of planner-predicted actions from 0 (equivalent to \textit{\cort OA}) to 1 (equivalent to \textit{\cort PA}).
% Results
The left side of Figure~\ref{fig:cms} shows that the smaller the fraction of oracle actions during training, the better the perplexity, up to an improvement of around 5\%.
Because perplexity evaluation happens with planner-predicted actions, the bigger the fraction of oracle actions during training, the bigger the training/evaluation mismatch, a problem known as exposure bias.

The perplexity improvement does not translate into improving generation metrics, with some metrics even consistently worsening.
To understand this, consider the plan-matching accuracy (green line).
As fewer actions are oracle, the plan-matching accuracy decreases, indicating the language model learns to rely less on the plan.
This suggests that generation metrics benefit from having a model rely more on the planner output, even if it is imperfect.
% Argue as avenue for future work

To try to get the best of both worlds, we also evaluate a setting with a scheduled fraction that linearly increases the fraction of planner-predicted actions from 0 to 1 during training, i.e., scheduled sampling, shown on the right side of Figure~\ref{fig:cms}.
However, we observe that this leads similar results as training only on planner-predicted actions.

Hence, overcoming this trade-off by both overcoming the problem of exposure bias and ensuring the language model learns to sufficiently rely on the proposed plans is an interesting avenue for future work.

\section{Conclusion}
Since end-to-end training is a key ingredient to the success of deep learning, it is important that we enable different system components to be optimized jointly.
In this work, we bridge the differentiability gap of a recent pretrained planning module with a language model by turning the indifferentiable hard-selection into a differentiable soft-selection. 
Our results demonstrate that this consistently improves perplexity.
We hope these findings can provide a foundation for enhancing production-scale language models through end-to-end planning mechanisms.

\section*{Acknowledgments}
This research was financed by the CALCULUS project—Commonsense and Anticipation enriched Learning of Continuous representations—European Research Council Advanced Grant H2020-ERC-2017-ADG 788506, \url{http://calculus-project.eu/}.

\clearpage
\section{Limitations}
% Authors are required to discuss the limitations of their work in a dedicated Section titled “Limitations”. This Section should be included at the end of the paper, before the references, and it will not count toward the page limit. This includes both, long and short papers. Papers without a limitations Section will be desk rejected. Note, prior to the December 2023 cycle, this was optional.
\subsection{Model Size}
Due to computational constraints, our evaluation was performed on relatively small models. Consequently, the scalability and effectiveness of the proposed method need to be validated on production-scale models to ensure its applicability in real-world scenarios.
\subsection{Planning Horizon}
Our approach involves planning only one step into the future. This is a simplification compared to how humans presumably think and plan farther into the future. Future work should investigate methods to extend the planning horizon, allowing the model to consider multiple future steps and thereby improve decision-making processes.

\subsubsection*{Broader Impact Statement}
% In this optional section, TMLR encourages authors to discuss possible repercussions of their work,
% notably any potential negative impact that a user of this research should be aware of. 
% Authors should consult the TMLR Ethics Guidelines available on the TMLR website
% for guidance on how to approach this subject.
While increasingly more capable LLMs are very useful, they can also be misused for harmful purposes (such as generating disinformation, helping in development of weapons, etc.). Because our work has used LLMs of modest size, there is little risk of it contributing to such misuses directly. It could do so however if our method would be used to make production-scale language models even more effective. If that is the case, it is important to take the necessary precautions before deployment, such as proper alignment with human values.

The compute requirements of large models also have a significant environmental impact~\citep{rillig2023risks}.
Use of a planning module also entails additional compute requirements, which can further contribute to this, although the planning module is relatively lightweight compared the the Language Models, and is invoked only once per sentence rather than for every token.

% in many journals. This is optional and at the discretion of the authors. Only add
% this information once your submission is accepted and deanonymized. 

% Only add this information once your submission is accepted and deanonymized. 

\bibliography{bib/manual,bib/zotero}
\bibliographystyle{tmlr}

\appendix

\section{Model Details}
\label{app:hyperparameters}
\paragraph{Parameter counts}
Table \ref{tab:model_details} shows parameter counts for our models

\begin{table}[ht]
\centering
\begin{tabular}{@{}l|r@{}}
\toprule
Model                         & Parameter Count   \\ \midrule
GPT2-Small                    & 124,439,808       \\ 
Olmo                          & 1,176,764,416     \\ 
Extra conditioning parameters & 13,770,240        \\
Planner parameters            & 116,378,496       \\ \bottomrule
\end{tabular}
\caption{Parameter counts for our models}
\label{tab:model_details}
\end{table}

\paragraph{Computational Budget}
We ran our experiments on either 12GB, 16GB or 24GB GPUs, each time using one GPU per experiment. We report a single run for each setting.
With this setting, joint fine-tuning of planner and LM takes around 40 hours for GPT2-Small and 60 hours for Olmo 1B.
Pretraining the planner for Next Action Prediction took around 90 hours, but we reuse the same pretrained planner for most experiments. Evaluating perplexity takes about 3-5 hours, while evaluating edit-distance (which requires generation) takes around 10-15 hours. 

We estimate that we ran about 100 experiments (only a subset of which led to results presented in the paper), which means in total we used around 7000 GPU hours.

\paragraph{Used artifacts}
% Our implementation extends upon the source code of \citet{cornille2024learning}, which was privately shared with us.
% Once their code is shared publicly, we will release our own extensions as soon as possible thereafter. Unless specified explicitly, all packages use default parameters.
We build on the source code of \citet{cornille2024learning}, which was shared with us privately. We will release our extensions publicly once their code is made available.

% Abstract writing actions are generated by first splitting every article into sentences using spaCy~\citep{honnibal2020spacy}, and then encoding them into embeddings using MPNet-base-v2~\citep{song_mpnet_2020} via the SentenceTransformer library~\citep{reimers2019sentence}\footnote{\url{https://huggingface.co/sentence-transformers/all-mpnet-base-v2}} to encode sentences into embeddings.
% The final clustering step is performed via Scikit-Learn~\citep{scikit-learn} with k-means++ initialization~\citep{arthur2007k}.
% All used libraries are either open source or freely usable for academic purposes.

We use spaCy~\citep{honnibal2020spacy} to split articles into sentences. These sentences are then transformed into embeddings using MPNet-base-v2~\citep{song_mpnet_2020} through the SentenceTransformer library~\citep{reimers2019sentence}\footnote{\url{https://huggingface.co/sentence-transformers/all-mpnet-base-v2}}. The final clustering is conducted using k-means++ initialization~\citep{arthur2007k} implemented in Scikit-Learn~\citep{scikit-learn}.

% We obtain the Wikipedia dataset through the `datasets` library at \url{https://huggingface.co/datasets/wikipedia} (version `20220301` from March 2022). No additional preprocessing is applied. We randomly subsample 285,310 articles for training, and 1,000 for each validation and test set, respectively. 

The Wikipedia dataset is accessed via the `datasets' library at \url{https://huggingface.co/datasets/wikipedia}, specifically the March 2022 version (`20220301`).

% The code base makes use of PyTorch~\citep{paszke2019pytorch}, the Huggingface `datasets`~\citep{lhoest-etal-2021-datasets} and `transformers`~\citep{wolf2020transformers} libraries to load and preprocess data and pretrained models (GPT-2~\citep{radford2019language}), respectively. Furthermore, we used PyTorch-Lightning~\citep{falcon2020pytorchlightning} for model training.

We use PyTorch~\citep{paszke2019pytorch}, the Huggingface `datasets'~\citep{lhoest2021datasets}, and `transformers'~\citep{wolf2020transformers} libraries for loading and preprocessing data and pretrained models (specifically GPT-2~\citep{radford2019language}). Additionally, we employ PyTorch-Lightning~\citep{falcon2020pytorchlightning} for model training.
All the libraries utilized are open source or freely available for academic use.

\paragraph{Hyperparameters}

Table \ref{tab:hyperparameters} shows hyperparameters used for our experiments.
We do not perform hyperparameter search for these, using the default hyperparameters reported in \cort.
\begin{table}[ht] % Adjust placement specifier if needed
\centering
\caption{Hyperparameter Settings} 
\label{tab:hyperparameters} 
\begin{tabular}{p{4cm}l} 
\toprule
\textbf{Hyperparameter} & \textbf{Value} \\
\midrule
Context window size      & 128 \\
Train $\mid$ test $\mid$ val split sizes & 285310 $\mid$ 1000 $\mid$ 1000 \\
% \multirow{3}{*}{Split sizes} 
%    & Train: 285310 \\ 
%    & Validate: 1000 \\
%    & Test: 1000 \\ 
K-means initialization & k-means++ \\
Default action count & 1024 \\
Action embedding dimension & 768 \\
\midrule

\multicolumn{2}{l}{\textbf{Language Model Fine-tuning}} \\ 
Batch size & 32 \\
Learning rate & 1e-4 \\ 
\midrule

\multicolumn{2}{l}{\textbf{Planner Training}} \\ 
Batch size & 32 \\
Learning rate & 1e-4 \\ 
\bottomrule
\end{tabular}
\end{table}

\section{Generation Evaluation Setup and Detailed Results}\label{app:geneval}
We follow the evaluation setup from \cort, and explain the details again here:

For MAUVE and Latent Perplexity, we generate 1024 tokens unconditionally (i.e., without context), matching the average length of the articles in the dataset. 

For ROUGE-2, and Edit distance, we use a prefix $t_1 \dots t_i$ from real texts and generate continuations from that prefix of 128, 256, 512, and 1024 tokens. 
Because Edit distance scales linearly with the number of tokens, we normalize the results across different lengths. For 128 tokens, we report the raw edit distance; for 256, we divide the edit distance by 2, and so on, ensuring a consistent comparison across generation lengths.

The results in the main text (Table \ref{tab:main_results}) are the average for these different generation lengths.

\end{document}

%% file: math_commands.tex
\usepackage{amsmath,amsfonts,bm}

\def\eqref#1{equation~\ref{#1}}
\def\1{\bm{1}}

\DeclareMathAlphabet{\mathsfit}{\encodingdefault}{\sfdefault}{m}{sl}
\SetMathAlphabet{\mathsfit}{bold}{\encodingdefault}{\sfdefault}{bx}{n}